\pdfoutput=1
\documentclass[11pt]{article}

\usepackage{emnlp2021}

\usepackage{todonotes}

\usepackage{times}
\usepackage{latexsym}
\usepackage{tabularx}
\usepackage{booktabs}

\usepackage[T1]{fontenc}
\usepackage[utf8]{inputenc}

\usepackage{microtype}
\usepackage{graphicx}
\usepackage{listings}
\usepackage{xcolor}
\usepackage{lipsum}
\definecolor{codegreen}{rgb}{0,0.6,0}
\definecolor{codegray}{rgb}{0.5,0.5,0.5}
\definecolor{codepurple}{rgb}{0.58,0,0.82}
\definecolor{backcolour}{rgb}{0.95,0.95,0.92}

\lstdefinestyle{mystyle}{
    backgroundcolor=\color{backcolour},   
    commentstyle=\color{codegreen},
    keywordstyle=\color{magenta},
    numberstyle=\tiny\color{codegray},
    numbers=none,
    stringstyle=\color{codepurple},
    basicstyle=\ttfamily\footnotesize,
    breakatwhitespace=false,         
    breaklines=false,                 
    captionpos=b,                    
    keepspaces=true,                 
    numbers=left,                    
    numbersep=5pt,                  
    showspaces=false,                
    showstringspaces=false,
    showtabs=false,                  
    tabsize=2
}
\DeclareRobustCommand{\huggingface}{%
  \begingroup\normalfont
  \vspace{-0.2em}%
  \raisebox{-0.4em}{%
  \includegraphics[height=1.5em]{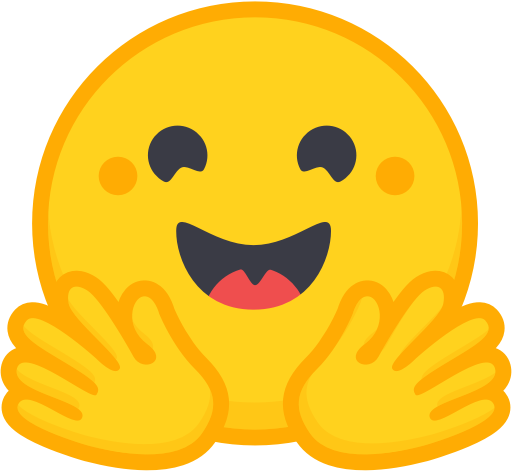}%
  }%
  \kern 0.4em%
  \endgroup
}
\title{\huggingface Datasets: A Community Library for Natural Language Processing}

 \author{ \vspace{-0.1cm} \\

 Quentin~Lhoest\thanks{\ \ Lead Library Maintainers, $^\Omega$ Library Creator, $^\uparrow$ Independent Research Contributor}~, Albert~Villanova~del~Moral$^*$,
                  Yacine Jernite,
                  Abhishek Thakur, \\
                  Patrick von Platen, 
                  Suraj Patil, 
                  Julien Chaumond, 
                  Mariama Drame,
                  Julien Plu,
                  Lewis Tunstall, \\
                  Joe Davison,
                  Mario Šaško$^\uparrow$,
                  Gunjan Chhablani$^\uparrow$,
                  Bhavitvya Malik$^\uparrow$,
                  Simon Brandeis, \\ 
                  Teven Le Scao,  
                  Victor Sanh, 
                  Canwen Xu, 
                  Nicolas Patry,
                  Angelina McMillan-Major, \\
                  Philipp Schmid,
                  Sylvain Gugger,
                  Clément Delangue,  
                  Théo Matussière,
                  Lysandre Debut, \\
                  Stas Bekman,
                  Pierric Cistac, 
                  Thibault Goehringer, 
                  Victor Mustar, 
                  François Lagunas, \\
                  Alexander M. Rush, and  Thomas~Wolf$\ ^\Omega$\\

\\
{ Hugging Face / \{quentin,thomas\}@huggingface.co }
}
\date{}

\begin{document}
\maketitle

\begin{abstract}
  The scale, variety, and quantity of publicly-available NLP
  datasets has grown rapidly as researchers propose new tasks, larger
  models, and novel benchmarks. \textit{Datasets} is a community
  library for contemporary NLP designed to support this ecosystem. \textit{Datasets} aims to standardize end-user interfaces,
  versioning, and documentation, while providing a lightweight
  front-end that behaves similarly for small datasets as for internet-scale
  corpora. The design of the library incorporates a distributed,
  community-driven approach to adding datasets and documenting
  usage. After a year of development, the library now includes more
  than 650 unique datasets, has more than 250 contributors, and has helped
  support a variety of novel cross-dataset research projects and shared tasks. The library is available at 
  \url{https://github.com/huggingface/datasets}.

\end{abstract}

\section{Introduction}

% Importance of datasets in NLP
% \footnotetext{$^*$ Lead Library Maintainers, $^\Omega$ Library Creator}
Datasets are central to empirical NLP: curated datasets 
are used for evaluation and benchmarks; supervised datasets are
used to train and fine-tune models; and large
unsupervised datasets are necessary for pretraining and language
modeling. Each dataset type differs in scale,
granularity and structure, in addition to  annotation methodology. 
Historically, new dataset paradigms have been crucial for driving
the development of NLP, from the Hansard corpus for
statistical machine translation~\cite{brown-etal-1988-statistical} to
the Penn Treebank for syntactic
modeling~\cite{marcus-etal-1993-building} to projects like OPUS and
Universal Dependencies~\cite{nivre-etal-2016-universal,tiedemann-nygaard-2004-opus}
which bring together cross-lingual data and annotations.

Contemporary NLP systems are now developed with a pipeline
that utilizes many different datasets at significantly varying
scale and level of annotation~\cite{peters-etal-2018-deep}. Different datasets are used for
pretraining, fine-tuning, and
benchmarking. As such, there has been a large increase in the number
of datasets utilized in the NLP community.  These include both large
text collections like C4~\cite{DBLP:journals/jmlr/RaffelSRLNMZLL20}, fine-tuning datasets like
SQuAD~\cite{rajpurkar-etal-2016-squad}, and even complex zero-shot challenge tasks. Benchmark datasets like GLUE have been central to quantifying the
the advances of models such as BERT~\cite{wang-etal-2018-glue, devlin-etal-2019-bert}.

% The rise of
% crowd-sourcing as a tool for soliciting annotations has made it
% possible to collect large annotation datasets from non-experts across
% a variety of different tasks. Additionally the influence of ImageNet
% benchmarking in computer vision~\cite{} has motivated the creation of
% new benchmark datasets in NLP such as SQuAD~\cite{}. This approach has
% already yielded some notable successes with the GLUE benchmark being
% the key first result demonstrating the gains from BERT~\cite{}.

% History

% Historically, new datasets have been central in driving the
% development NLP has been a field. Early efforts in language
% modeling~\cite{} were made possible through availability of corpora
% like Reuters. Similarly large corpora like the Hansard corpus~\cite{}
% were critical to the developments of statistical methods for
% translation~\cite{}. The Penn Treebank project showed that careful and
% precise annotation of linguistic structure made it possible to learn
% complex syntactic models~\cite{}. These English-centric projects have
% been followed by many other efforts such as Opus or the Universal
% Dependency Project which extend similar resources to many different languages~\cite{}.

% Growth in datasets

% Standardization and Versioning. 

The growth in datasets also brings significant challenges, including interface standardization, versioning, and documentation.
A practitioner should be able to utilize $N$ different datasets without requiring $N$ different interfaces.
In addition, $N$ practitioners using the same dataset should know they have exactly the same version.
Datasets have also grown larger, and ideally interfaces should not have to change due to this scale, whether one is using small-scale datasets like Climate Fever ($\sim$1k data points), medium-scale Yahoo Answers ($\sim$1M), or even all of PubMed ($\sim$79B)\nocite{diggelmann2020climatefever}.
Finally, datasets are being created with a variety of different procedures, from crowd-sourcing to scraping to synthetic generation, which need to be taken into account when evaluating which is most appropriate for a given purpose and ought to be immediately apparent to prospective users \cite{DBLP:journals/corr/abs-1803-09010}.

% Library for collecting and integrating datasets

\textit{Datasets} is a community library designed to address the challenges of dataset
management and access, while supporting community culture and
norms. The library targets the following goals:

\begin{itemize}
\item Ease-of-use and Standardization: All datasets can be easily
  downloaded with one line of code. Each dataset utilizes a standard
  tabular format, and is versioned and cited.
\item Efficiency and Scale: Datasets are computation- and memory-efficient
  by default and work seamlessly with tokenization and featurization.
  Massive datasets can even be streamed through the same interface.
\item Community and Documentation: The project is
  community-built and has hundreds of contributors across
  languages. Each dataset is tagged and documented with a
  datasheet describing its usage, types, and construction.
\end{itemize}

\textit{Datasets} is in continual development by the engineers at Hugging Face and is released under an Apache 2.0 license.\footnote{Datasets themselves may utilize different licenses which are documented in the library.} The library is available at 
  \url{https://github.com/huggingface/datasets}. Full documentation is available through the project website.\footnote{\url{https://huggingface.co/docs/datasets/}}

\section{Related Work}

There is a long history of projects aiming to group,
categorize, version, and distribute NLP datasets which we briefly survey. Most notably, the
Linguistic Data Consortium (LDC) stores, serves, and manages a variety
of datasets for language and speech.  In addition to hosting and
distributing corpus resources, the LDC supports
significant annotation efforts.  Other projects have aimed to collect
related annotations together.  Projects like OntoNotes have collected
annotations across multiple tasks for a single
corpus~\cite{pradhan-xue-2009-ontonotes} whereas the Universal
Dependency treebank~\cite{nivre-etal-2016-universal} collects similar
annotations across languages. In machine translation, projects like
OPUS catalog the translation resources for many different
languages. These differ from \textit{Datasets} which collects and
provides access to datasets in a content-agnostic way.

Other projects have aimed to make it easy to access core NLP datasets.
The influential NLTK project~\cite{bird-2006-nltk} provided a data
library that makes it easy to download and access core datasets. SpaCy also provides a similar loading
interface~\cite{honnibal2017spacy}.  In recent years, concurrent with
the move towards deep learning, there has been a growth in large
freely available datasets often with less precise annotation
standards. This has motivated cloud-based repositories of
datasets. Initiatives like \citet{TFDS} and \citet{TorchText} have collected various datasets in a common cloud
format. This project began as a fork of TensorFlow-Datasets,
but has diverged significantly. 

\textit{Datasets} differs from these projects along several axes. 
The project is decoupled from any modeling framework and provides a general-purpose tabular API. 
It focuses on NLP specifically and provides specialized types and structures for language constructs. 
Finally, it prioritizes community management and documentation through the dataset hub and data cards, 
and aims to provide access to a long-tail of datasets for many tasks and languages. 

% Torch NLP

% \begin{figure*}
%   \centering
%   \missingfigure{A graph of the usage of the library}
%   \caption{Usage graphic}
% \end{figure*}

\section{Library Tour and Design}
\label{sec:library_tour}

We begin with a brief tour. Accessing a dataset is done simply by referring to it by a global identity.  

\begin{lstlisting}[language=python,numbers=none]
dataset = load_dataset("boolq")
\end{lstlisting}

\noindent Each dataset has a features schema and  metadata.

\begin{lstlisting}[language=python,numbers=none]
print(dataset.features, dataset.info)
\end{lstlisting}

\noindent Any slice of data points can be accessed directly without loading the 
full dataset into memory.

\begin{lstlisting}[language=python,numbers=none]
dataset["train"][start:end]
\end{lstlisting}

\noindent  Processing can be applied to every data point in a batched and parallel 
   fashion using standard libraries such as NumPy or Torch.

\begin{lstlisting}[language=python,numbers=none]
# Torch function "tokenize" 
tokenized = dataset.map(tokenize, 
                        num_proc=32)
\end{lstlisting}

\noindent \textit{Datasets} facilitates each of these four Stages with the following technical steps.

\paragraph{S1. Dataset Retrieval and Building}  \textit{Datasets} does not host the underlying raw datasets, but accesses hosted data from the original authors in a distributed manner.\footnote{For datasets with intensive preprocessing, such as Wikipedia, a preprocessed version is hosted. Datasets removed by the author are not centrally cached and become unavailable.} Each dataset has a community contributed builder module. The builder module has the responsibility of processing the raw data, e.g. text or CSV, into a common dataset interface representation.

\paragraph{S2. Data Point Representation} Each built dataset is represented internally as a table with typed columns. The \textit{Dataset} type system includes a variety of common and NLP-targeted types. In addition to atomic values (int's, float's, string's, binary blobs) and JSON-like dicts and lists, the library also includes named categorical class labels, sequences, paired translations, and higher-dimension arrays for images, videos, or waveforms. 

\paragraph{S3. In-Memory Access} \textit{Datasets} is built on top of Apache Arrow, a cross-language columnar data framework~\cite{ApacheArrow}. Arrow provides a local caching system allowing datasets to be backed by an on-disk cache, which is memory-mapped for fast lookup. This architecture allows for large datasets to be used on machines with relatively small device memory. Arrow also allows for copy-free hand-offs to standard machine learning tools such as NumPy, Pandas, Torch, and TensorFlow.

\paragraph{S4. User Processing} At download, the library provides access to the typed data with minimal preprocessing. It provides functions for dataset manipulation including sorting, shuffling, splitting, and filtering. For complex manipulations, it provides a powerful \textit{map} function that supports arbitrary Python functions for creating new in-memory tables. For large datasets, map can be run in batched, multi-process mode to apply processing in parallel. Furthermore, data processed by the same function is automatically cached between sessions.

\paragraph{Complete Flow} Upon requesting a dataset, it is downloaded from the original host. This triggers dataset-specific builder code which converts the text into a typed tabular format matching the feature schema and caches the table. The user is given a memory-mapped typed table. To further process the data, e.g. tokenize, the user can run arbitrary vectorized code and cache the results. 

\section{Dataset Documentation and Search}

\begin{figure}
  \centering
  \includegraphics[width=\linewidth]{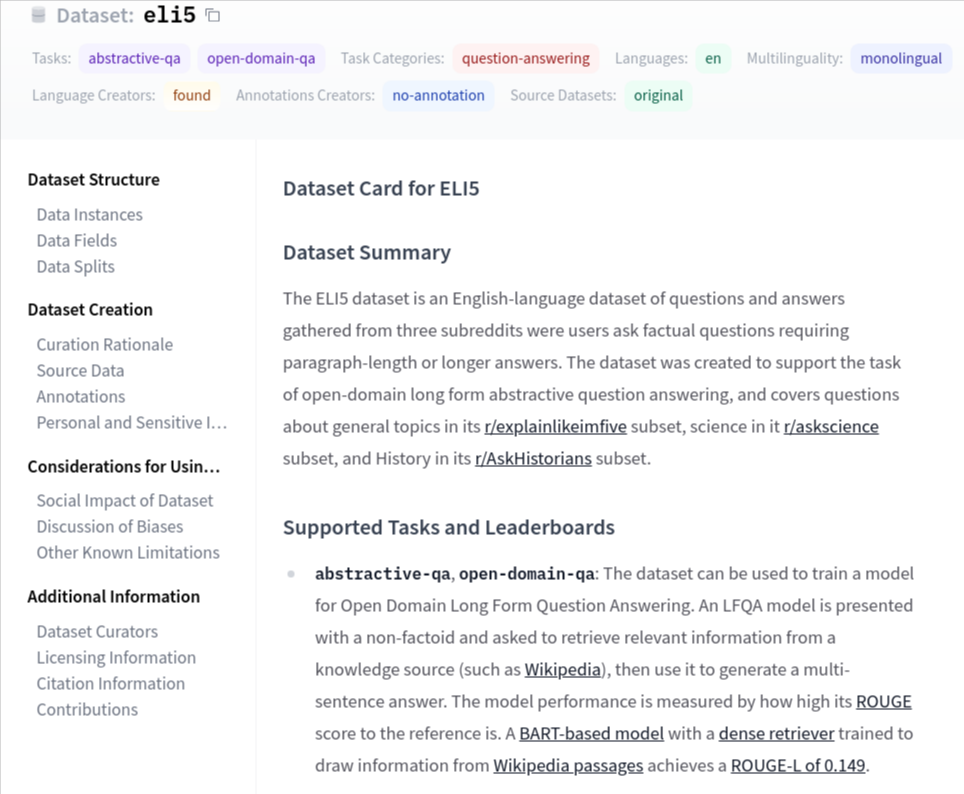}
  \caption{The data card for ELI5~\cite{fan-etal-2019-eli5}.}
\label{fig:datasheet}
\end{figure}

%Section~\ref{sec:library_tour} describes how the core of the \textit{Datasets} library meets the efficiency requirements we identified for utilizing a given dataset. Additionally, 

\textit{Datasets} is backed by the \textit{Dataset Hub}~\footnote{\url{https://hf.co/datasets/}} that helps users navigate the growing number of available resources and draws inspiration from recent work calling for better documentation of ML datasets in general~\cite{DBLP:journals/corr/abs-1803-09010} and NLP datasets in particular~\cite{data_statements}.

Datasets can be seen as a form of infrastructure~\cite{data_infra}.
NLP practitioners typically make use of them with a specific goal in mind, whether they are looking to answer a specified research question or developing a system for a particular practical application.
To that end, they need to be able to not only easily identify which dataset is most appropriate for the task at hand, but also to understand how various properties of that best candidate might help with, or, conversely, run contrary to their purpose.

The \textit{Dataset Hub} includes all of the datasets available in the library. It links each of them together though: a set of \textit{structured tags} holding information about their languages, tasks supported, licenses, etc.; a \textit{data card} based on a template\footnote{\url{https://hf.co/datasets/card-guide}} designed to combine relevant technical considerations and broader context information~\cite{McMillanMajor2021cards}; and a \textit{list of models} trained on the dataset. Both the tags and data card are filled manually by the contributor who introduces the dataset to the library. Figure~\ref{fig:datasheet} presents an example of the dataset page on the hub.\footnote{\url{https://hf.co/datasets/eli5}} Together, these pages and the search interface help users navigate the available resources.

\paragraph{Choosing a Dataset} Given a use case, the structured tags provide a way to surface helpful datasets. For example, requesting all datasets that have the tags for Spanish language and the Question Answering task category returns 7 items at the time of writing. A user can then refine their choice by reading through the data cards, which contain sections describing the variety of language used, legal considerations including licensing and incidence of Personal Identifying Information, and paragraphs about known social biases resulting from the collection process that might lead a deployed model to cause disparate harms.

\paragraph{Using a Dataset} The data card also contains information to help users navigate all the choices, from hardware to modeling, that go into successfully training a system. These include the number of examples in each of the dataset splits, the size on disk of the data, meaningful differences between the training, validation, and test split, and free text descriptions of the various fields that make up each example to help decide what information to use as input or output of a prediction model.

\paragraph{The Data Card as a Living Document} A dataset's life continues beyond its initial release. As NLP practitioners interact with the dataset in various ways, they may surface annotation artifacts that affect the behavior of trained models in unexpected ways~\cite{snli_artifacts},\footnote{\url{https://hf.co/datasets/snli\#other-known-limitations}} issues in the way the standard split was initially devised to test a model's ability to adapt to new settings~\cite{lfqa_hurdles}, or new understanding of the social biases exhibited therein~\cite{barrier_disability}. The community-driven nature of \textit{Datasets} and the versioning mechanisms provided by the GitHub backend provide an opportunity to keep the data cards up to date as information comes to light and to make gradual progress toward having as complete documentation as possible.

% \begin{table}
%   \centering
%  \begin{tabularx}{230px}{lX}
%   \toprule
%  Desc.& Summary, Leaderboard, Languages\\
%  Structure& Data Instances, Fields, Splits\\
%  Creation& Curation, Source, Annotations,  Sensitive information\\
%  Impact & Social Impact, Biases,  Known Limitations\\
%  Additional & Licenses, Citations \\
%   \bottomrule
% \end{tabularx}
%   \label{tab:dprop}
% \end{table}

\section{Dataset Usage and Use-Cases}
\label{sec:build}

\begin{figure*}
  \centering
  \begin{minipage}{0.45\linewidth}
  \includegraphics[width=\linewidth]{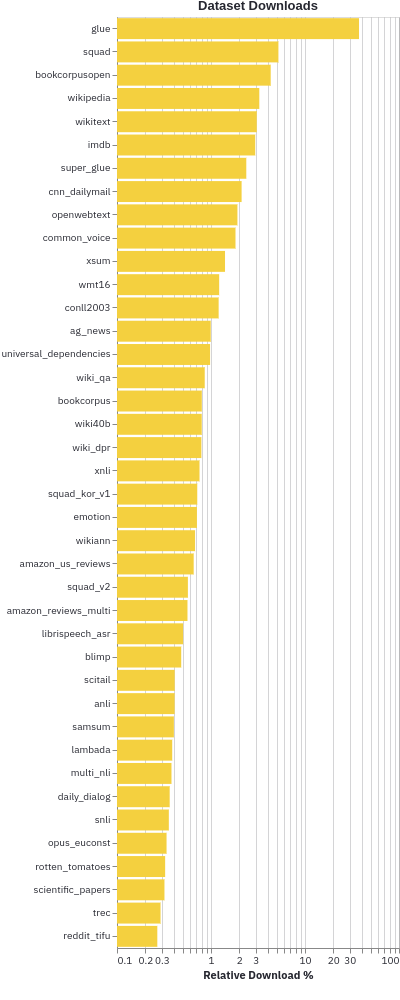} 

  \end{minipage} \hspace*{0.5cm}\begin{minipage}{0.45\linewidth}
   \includegraphics[width=\linewidth]{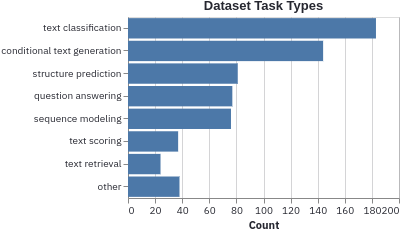}
\vspace{0.25cm}

   \includegraphics[width=\linewidth]{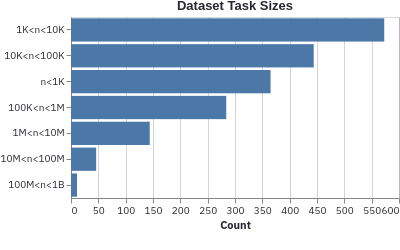}
\vspace{0.25cm}

   \includegraphics[width=\linewidth]{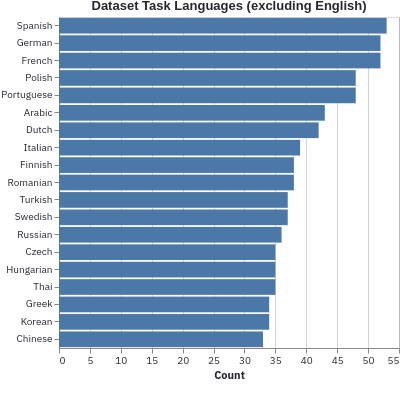}

  \end{minipage} 
  \caption{Summary statistics from the datasets in the library. (\textbf{Left})
    The relative download numbers of the most popular datasets
    in the library. (\textbf{Right}) Task properties. Each dataset may have multiple sub-tasks. Task Types are the types labeled in the library. Task Sizes are the number of data points in the table. Task Languages are the languages tagged in the library (many datasets 
    include tasks in different languages).
    \label{fig:datasets}
   }
\end{figure*}

\textit{Datasets} is now being actively used for a variety of tasks.
Figure~\ref{fig:datasets}~(left) shows statistics about library usage. We can see that the most commonly downloaded libraries are 
popular English benchmarks such as GLUE and SQuAD which are often used for teaching and examples.
However there is a range of popular models for different tasks and languages.

Figure~\ref{fig:datasets}~(right) shows the wide coverage of the library in terms of task types, sizes, and languages, with 
currently 681 total datasets.
During the development of the \textit{Datasets} project, there was a public hackathon to have community members develop new Dataset builders and add them to the project. This event led 
485 commits and 285 unique contributors to the library. Recent work has outlined the difficulty of finding data sources for lower-resourced languages through automatic filtering alone~\cite{data_audit}. The breadth of languages spoken by participants in this event made it possible to more reliably bootstrap the library with datasets in a wide range of different languages. Finally while \textit{Datasets} is designed for NLP, it is becoming used for multi-modal datasets. The library now includes types for continuous data, including multi-dimensional arrays for image and video data and an \textit{Audio} type.

% This work helped bring in many datasets outside the specific of the NLP research community.
%These formats are utilized to hosts many of the core image recognition datasets such as MNIST~\cite{lecun2010mnist}, as well as speech datasets such as Librispeech~\cite{panayotov2015librispeech}. 

% \textcolor{red}{{TODO Yacine} -Challenge: in spite of recent calls to provide much needed documentation, adoption of a general schema has been slow - number of already available language dataset makes \textit{post hoc} annotation and unified documentation difficult for any central entity~\cite{McMillanMajor2021cards}. - but one person contributes at a time - validated by CI - choice to leave [Needs More Information]}

% To encourage adding more datasets, \textit{Datasets} includes a simple
% builder interface based off of the builders in TensorFlow Datasets.
% Adding a new dataset to the library consists of implementing a Builder
% subclass. The class needs to download the raw data and process it into
% a typed Apache Arrow table. In addition, the class has the
% responsibility of declaring the metadata for the dataset including a
% description, citation to the original paper and website, table type,
% languages, and license. To further ease the use of the builder
% interfaces, there are a collection of standard builders from popular
% formats including CSV, JSON, and Pandas data, which can be used for public datasets
% or for private builders.

\subsection{Case Studies: $N$-Dataset NLP}

A standardized library of datasets opens up new use-cases
beyond making single datasets easy to download. We highlight three
use-cases in which practitioners have employed the \textit{Datasets} library.

\paragraph{Case Study 1: $N$-task Pretraining Benchmarks}

Benchmarking frameworks such as NLP Decathlon and GLUE have
popularized the comparison of a single NLP model across a
variety of tasks~\cite{McCann2018TheNL,wang-etal-2018-glue}.  Recently benchmarking frameworks
like GPT-3's test suite framework~\cite{DBLP:conf/nips/BrownMRSKDNSSAA20} have expanded this
benchmarking style even further, taking on dozens of
different tasks. This research has increased interest in
 comparison of different datasets at scale.

\textit{Datasets} is designed to facilitate large-scale, $N$-task
benchmarking beyond what might be possible for a single researcher to
set up. For example, the Eleuther AI project aims to produce a massive
scale open-source model. As part of
this project they have released an \textit{LM Evaluation
Harness}\footnote{https://github.com/EleutherAI/lm-evaluation-harness}
which includes nearly 100 different NLP tasks to test a large scale
language model. This framework is built with the
\textit{Datasets} library as a method for retrieving and caching
datasets.

\paragraph{Case Study 2: Reproducible Shared Tasks}

NLP has a tradition of shared tasks that become long-lived benchmark
datasets. Tasks like  CoNLL 2000~\cite{tjong-kim-sang-buchholz-2000-introduction} continue
to be widely used more than 20 years after their
release. \textit{Datasets} provides a convenient,
reproducible, and standardized method for hosting and maintaining
shared tasks, particularly when they require multiple different
datasets. 

\textit{Datasets} was used to support the first GEM
(Generation, Evaluation, and Metrics) workshop~\cite{Gehrmann2021TheGB}.  This workshop
ran a shared task comparing natural language generation (NLG) systems
on 12 different tasks. The tasks included examples from twenty
different languages and supervised datasets varying from size of 5k
examples to 500k. Critically, the shared task had a large variety of
different input formats including tables, articles, RDF triples, and
meaning graphs. \textit{Datasets} allows users to access all
12 datasets with a single line of code in their shared task
description. 
%This code will continue to run and work indefinitely.

\paragraph{Case Study 3: Robustness Evaluation}

While NLP models have improved to the point that on-paper they compete
with human performance, many research projects have demonstrated that
these same models are fooled when given out-of-domain
examples~\cite{koehn2017six}, simple adversarial constructions~\cite{belinkov2018synthetic}, or
examples that spuriously match basic patterns~\cite{poliak-etal-2018-hypothesis}.
 
\textit{Datasets} can be used to support better benchmarking of these issues. The \textit{Robustness Gym}\footnote{https://robustnessgym.com/} proposes a systematic way to test an NLP system across many different proposed techniques, specifically subpopulations, transformations, evaluation sets, and adversarial attacks~\cite{goel-etal-2021-robustness}. Together, these provide a robustness report that is more specific than a single evaluation measure. While developed independently, the Robustness Gym is built on \textit{Datasets}, and "relies on a common data interface" provided by the library.

% to include these additional
% datasets within standard model comparison. By standardizing loading
% and model interface, a researcher does not need to do any additional
% work to include a wide set of different methods and models within
% their experimental results. \cite{victor} Furthermore the metadata and
% descriptions make it easy to group together these auxiliary datasets
% with the original task as well as codigin the specific generation
% procedure for which they were constructed.

\section{Additional Functionality and Uses}
\noindent \textbf{Streaming}
Some datasets are extremely large and cannot even fit on disk.
\textit{Datasets} includes a streaming mode that buffers these datasets on the fly. 
This mode supports the core map primitive, which works on each data batch as it is streamed. 
Datasets streaming helped enable recent research into distributed training of a very large open NLP  model~\cite{diskin2021distributed}.

\noindent\textbf{Indexing}
 \textit{Datasets} includes tools for easily building and utilizing a search index over an arbitrary dataset. To construct the index the library can interface either with FAISS or ElasticSearch~\cite{JDH17,ELS}. This interface makes it easy to efficiently find nearest neighbors either with textual or vector queries. Indexing was used to host the open-source version of Retrieval-Augmented Generation~\cite{DBLP:journals/corr/abs-2005-11401}, a generation model backed by the ability to query knowledge from large-scale knowledge sources.

\noindent \textbf{Metrics}
\textit{Datasets} includes an interface for standardizing \textit{metrics} which can be documented, versioned and  matched with datasets. This functionality is particularly useful for benchmark datasets such as GLUE that include 
multiple tasks each with their own metric. Some metrics like
BLEU and SQuAD are included directly in the library code, whereas others
are linked to external packages. The library also allows for metrics to
be applied in a distributed manner over the
dataset.

\begin{figure}
  \centering
  \includegraphics[width=\linewidth]{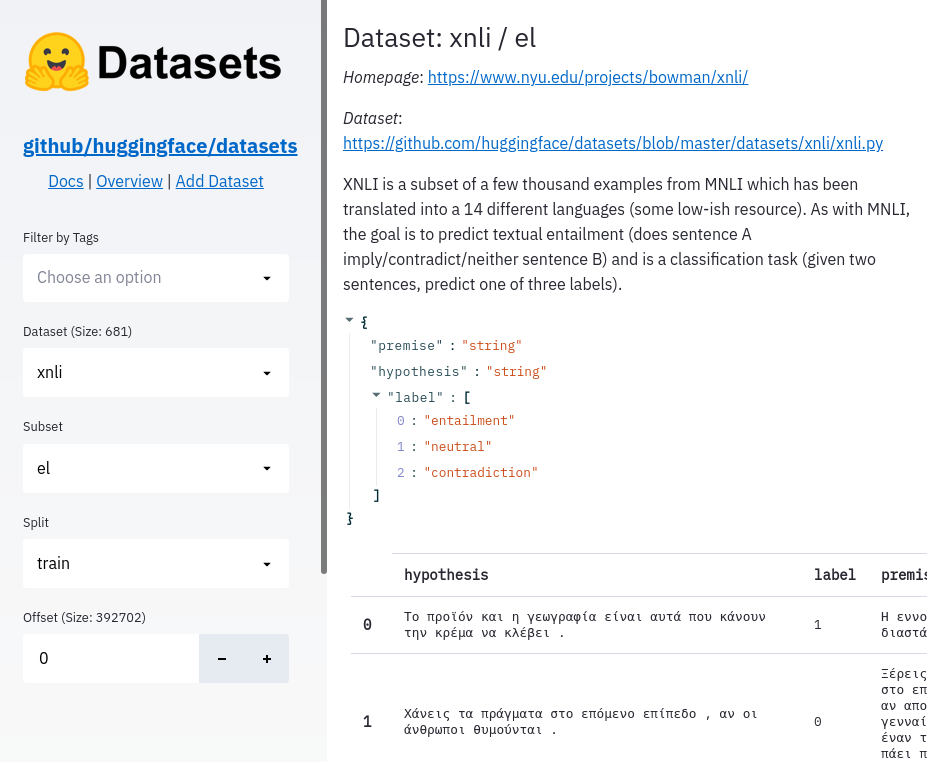}
  \caption{\textit{Datasets} viewer is an application that 
  shows all rows for all datasets in the library. The interface 
  allows users to change datasets, subsets, and splits, while seeing the 
  dataset schema and metadata.\label{fig:view}}
\end{figure}

\noindent\textbf{Data Viewer}
A benefit of the standardized interface of the library is that it
makes it trivial to build a cross-task dataset viewer. As an example,
Hugging Face hosts a generic viewer for the entirety of
\textit{datasets} (Figure~\ref{fig:view}) \footnote{https://huggingface.co/datasets/viewer/}.
In this viewer, anyone on the web can open all almost 650 different
datasets and view any example. Because the tables are
typed, the viewer can easily show all component features, structured data,
and multi-modal features.

\section{Conclusion}

Hugging Face \textit{Datasets} is an open-source, community-driven
library that standardizes the processing, distribution, and
documentation of NLP datasets. The core library is designed to be 
easy to use, fast, and to use the same interface for datasets 
of varying size. At 650 datasets from over 250
contributors, it makes it easy to use standard datasets, has facilitated new use cases of cross-dataset NLP, 
and has advanced features for tasks like indexing and streaming large datasets.

\pagebreak

\section*{Acknowledgements}

While organized by Hugging Face, \textit{Datasets} is an open-source project driven by contributors. This work was only possible thanks to Charin Polpanumas, Cahya Wirawan, Jonatas Grosman, Thomas Hudson, Zaid Alyafeai, Rahul Chauhan, Vineeth S, Sandip, Yvonnegitau, Jared T Nielsen, Michal Jamry, Bharat Raghunathan, Ceyda Cinarel, David Adelani, Misbah Khan, Steven Liu, Vasudev Gupta, Matthew Bui, Abdul Rafay Khalid, Beth Tenorio, Eduardo Gonzalez Ponferrada, Harshal Mittal, Hugo Abonizio, Moussa Kamal Eddine, Stefan Schweter, Sumanth Doddapaneni, Yavuz Kömeçoğlu, Yusuke Mori, J-chim, Ontocord, Skyprince999, Vrindaprabhu, Jonathan Bragg, Philip May, Alexander Seifert, Ivanzidov, Jake Tae, Karim Foda, Mohamed Al Salti, Nick Doiron, Vinay, Czabo, Vblagoje, Nilansh Rajput, Abdulelah S. Al Mesfer, Akshay Bhardwaj, Amit Moryossef, Basava Sai Naga Viswa Chaitanya, Darek Kłeczek, Darshan Gandhi, Gustavo Aguilar, Hassan Ismail Fawaz, Jack Morris, Jamesg, Jonathan Chang, Karthik Bhaskar, Manan Dey, Maria Grandury, Michael A. Hedderich, Mounica Maddela, Nathan Cooper, Purvi M, Richard Wang, Song Feng, Sourab Mangrulkar, Tanmoy, Vijayasaradhi, Zacharysbrown, Chameleontk, Eusip, Jeromeku, Patpizio, Tuner007, Benjamin Van Der Burgh, Bharati Patidar, George Mihaila, Olivier, Tim Isbister, Alessandro Suglia, Başak Buluz Kömeçoğlu, Boris Dayma, Dariusz Kajtoch, Frankie Robertson, Jieyu, Mihaelagaman, Nikhil Bartwal, Param Bhavsar, Paullerner, Rachelker, Ricardo Rei, Sai, Sasha Rush, Suraj Parmar, Takuro Niitsuma, Taycir Yahmed, Tuan-phong Nguyen, Vladimir Gurevich, Alex, Calpt, Idoh, Justin-yan, Katnoria, Sileod, Avinash Swaminathan, Connor Mccarthy, Jungwhan Kim, Leo Zhao, Sanjay Kamath, (bill) Yuchen Lin, 2dot71mily, 8bitmp3, Abi Komma, Adam, Adeep Hande, Aditya Siddhant, Akash Kumar Gautam, Alaa Houimel, Alex Dong, Along, Anastasia Shimorina, Andre Barbosa, Anton Lozhkov, Antonio V Mendoza, Ashmeet Lamba, Ayushi Dalmia, Batjedi, Behçet Şentürk, Bernardt Duvenhage, Binny Mathew, Birger Moëll, Blanc Ray, Bram Vanroy, Clément Rebuffel, Daniel Khashabi, David Fidalgo, David Wadden, Dhruv Kumar, Diwakar Mahajan, Elron Bandel, Emrah Budur, Fatima Haouari, Fraser Greenlee, Gergely Nemeth, Gowtham.r, Hemil Desai, Hiroki Nakayama, Ilham F Putra, Jannis Vamvas, Javier De La Rosa, Javier-jimenez99, Jeff Hale, Jeff Yang, Joel Niklaus, John Miller, John Mollas, Joshua Adelman, Juan Julián Cea Morán, Kacper Łukawski, Koichi Miyamoto, Kushal Kedia, Laxya Agarwal, Leandro Von Werra, Loïc Estève, Luca Di Liello, Malik Altakrori, Manuel, Maramhasanain, Marcin Flis, Matteo Manica, Matthew Peters, Mehrdad Farahani, Merve Noyan, Mihai Ilie, Mitchell Gordon, Niccolò Campolungo, Nihal Harish, Noa Onoszko, Nora Belrose, Or Sharir, Oyvind Tafjord, Pewolf, Pariente Manuel, Pasquale Minervini, Pedro Ortiz Suárez, Pedro Lima, Pengcheng Yin, Petros Stavropoulos, Phil Wang, Philipp Christmann, Philipp Dufter, Philippe Laban, Pierre Colombo, Rahul Danu, Rabeeh Karimi Mahabadi, Remi Calizzano, Reshinth Adithyan, Rodion Martynov, Roman Tezikov, Sam Shleifer, Savaş Yıldırım, Sergey Mkrtchyan, Shubham Jain, Shubhambindal2017, Subhendu Ranjan Mishra, Taimur Ibrahim, Tanmay Thakur, Thomas Diggelmann, Théophile Blard, Tobias Slott, Tsvetomila Mihaylova, Vaibhav Adlakha, Vegar Andreas Bergum, Victor Velev, Vlad Lialin, Wilson Lee, Yang Wang, Yasir Abdurrohman, Yenting (Adam) Lin, Yixin Nie, Yoav Artzi, Yoni Gottesman, Yongrae Jo, Yuxiang Wu, Zhong Peixiang, Zihan Wang, Aditya2211, Alejandrocros, Andy Zou, Brainshawn, Cemilcengiz, Chutaklee, Gaurav Rai, Dhruvjoshi1998, Duttahritwik, Enod, Felixgwu, Ggdupont, Jerryishere, Jeswan, Lodgi, Lorinczb, Maxbartolo, Nathan Dahlberg, Neal, Ngdodd, Kristo, Onur Güngör, Ophelielacroix, Padipadou, and Phiwi.

% Entries for the entire Anthology, followed by custom entries
\bibliography{anthology,custom}
\bibliographystyle{acl_natbib}

\end{document}